\documentclass[preprint,11pt]{elsarticle}
\usepackage[margin=1in]{geometry} 

\usepackage{graphicx}%
\usepackage{multirow}%
\usepackage{amsmath,amssymb,amsfonts}%
\usepackage{amsthm}%
\usepackage{mathrsfs}%
\usepackage[title]{appendix}%
\usepackage{xcolor}%
\usepackage{textcomp}%
\usepackage{manyfoot}%
\usepackage{booktabs}%
\usepackage{algorithm}%
\usepackage{algorithmicx}%
\usepackage{algpseudocode}%
\usepackage{listings}%
\usepackage{array}
\usepackage{subcaption}
\usepackage{booktabs}
\begin{document}

\begin{frontmatter}



\title{Semantic-Driven Topic Modeling Using Transformer-Based Embeddings and Clustering Algorithms}





\author[a]{Melkamu Abay Mersha } 
\author[b]{Mesay Gemeda yigezu }
\author[a]{Jugal Kalita}

\address[a]{Department of Computer Science, University of Colorado Colorado Springs (UCCS), Colorado Springs, USA}
\address[b]{Instituto Politécnico Nacional (IPN), Centro de Investigación en Computación (CIC), Mexico city, Mexico}

\begin{abstract}
Topic modeling is a powerful technique to discover hidden topics and patterns within a collection of documents without prior knowledge. Traditional topic modeling and clustering-based techniques encounter challenges in capturing contextual semantic information. This study introduces an innovative end-to-end semantic-driven topic modeling technique for the topic extraction process, utilizing advanced word and document embeddings combined with a powerful clustering algorithm. This semantic-driven approach represents a significant advancement in topic modeling methodologies. It leverages contextual semantic information to extract coherent and meaningful topics. Specifically, our model generates document embeddings using pre-trained transformer-based language models, reduces the dimensions of the embeddings, clusters the embeddings based on semantic similarity, and generates coherent topics for each cluster. Compared to ChatGPT and traditional topic modeling algorithms, our model provides more coherent and meaningful topics.
\end{abstract}

\begin{keyword}
Topic Modeling \sep Semantic \sep  Cluster \sep  Transformer-Based Embeddings \sep Transformer \sep Topic Extraction \sep Semantic-Driven \sep Deep Learning \sep Natural Language Processing. \\

\end{keyword}

\end{frontmatter}


\section{Introduction}
Topic modeling is a powerful technique used to discover hidden topics or latent thematic patterns within a collection of documents without prior knowledge \cite{blei2012probabilistic}. Topic modeling helps extract significant and meaningful topics from documents and provides valuable insights into the document's ideas. Topic modeling is essential in natural language processing and machine learning for reasons such as data exploration and understanding \cite{rodriguez2020computational}, document organization and summarization \cite{joshi2023deepsumm}, information retrieval \cite{albalawi2020using}, recommendation systems, content analysis \cite{punziano2023digital},  market research and customer insights \cite{mustak2021artificial}, and textual data preprocessing \cite{sbalchiero2020topic}. 

Traditional topic modeling methods such as Latent Dirichlet Allocation (LDA) \cite{blei2003latent}, Non-Negative Matrix Factorization (NMF) \cite{fevotte2011algorithms}, Latent Semantic Analysis (LSA) \cite{hofmann1999probabilistic}, and some BERT-based topic models work based on the bag-of-words approach to extract topics. Due to reliance on the bag-of-words technique, they suffer from the limitation that they treat all words in isolation without considering contextual relevance and relationships of words to the document. Traditional and even some Transformer-based topic models \cite{grootendorst2022bertopic} encounter challenges in contextual understanding at the topic extraction stage, potentially leading to less accurate and meaningful topic representations from the document collection.

In this study, we present a novel semantic-driven topic modeling approach that leverages the Transformer's ability to capture contextual information about words within the document throughout the end-to-end topic extraction process. We ensure that the model focuses only on the most relevant words within each document, disregarding non-relevant ones. This unique feature of our model sets it apart from others and enhances its ability to extract accurate and meaningful topics for each group of documents. We hypothesize that a unique word with no contextual relevance to the document is not a good topic representative for that document. This enables the proposed model to extract more accurate and meaningful topics for each group of documents. To the best of our knowledge, this semantic-driven end-to-end topic extraction approach is our innovative work.

Our model, designed with four layers, plays a pivotal role in utilizing the contextual information generated by Transformers for words and sentences from the given documents during topic extraction. This not only allows for a deeper understanding of documents but also significantly improves the quality of extracted topics. By combining these four layers and leveraging the power of Transformer's contextual embeddings, our model outperforms existing topic techniques such as LDA \cite{blei2003latent}, Embedded Topic Model (ETM) \cite{dieng2020topic}, Correlated Topic Model (CTM) \cite{blei2006correlated}, and BERTopic \cite{grootendorst2022bertopic}. Our work makes the following contributions.
\begin{itemize}
\item Developing a novel semantic-driven topic modeling technique for an end-to-end topic extraction process.
  \item We extract quality and coherent topics leveraging rich contextual information about word usage available within the document.
  \item We further improve the model's performance by eliminating non-relevant topic representative words in a second layer of processing once again based on the contextual information. 
\end{itemize}

The paper is organized as follows: a review of the most recent related works is presented in Section 2. The model architecture and functions of the components are discussed in Section 3. Section 4 covers the experimental setup, results, and analysis. Finally, the paper concludes with the findings in Section 5.

\section{Related Works}
The current state-of-the-art topic modeling methodologies can be classified into two main categories: probabilistic and embedding-based. Probabilistic models like LDA \cite{blei2003latent}, NMF \cite{fevotte2011algorithms}, LSA \cite{hofmann1999probabilistic}, and other variants of LDA work based on the statistical properties of data. However, these probabilistic models have a few limitations when using bag-of-words representation. The embedding-based models use text embeddings and can overcome the limitations of the traditional probabilistic-based models.

In recent years, topic modeling has shown improvement by exploiting the power of neural network models to enhance traditional techniques, resulting in improved performance and the ability to capture more complex relationships within large document collections \cite{zhao2021topic} and  \cite{terragni2021octis}.  The integration of word embeddings into classical probabilistic models has shown effective and promising topic representations \cite{agarwal2021comparative} and \cite{qiang2017topic}. There has been a substantial surge in the development of topic-modeling techniques, primarily focused on embedding-based models \cite{dieng2020topic, bianchi2020cross,yigezu2023habesha}. Embedding-based models have achieved good performance because of their capability to capture the contextual meaning and the semantic relationship among words in a document. Angelov (2020) introduced an advanced topic modeling approach that utilizes clusters of pre-trained word embeddings instead of traditional probabilistic topic model methods \cite{angelov2020top2vec}. The authors achieved faster and more efficient topic extraction, generating promising results with accurate topics for each cluster.   Bianchi et al. (2020) also demonstrated the utilization of word embeddings to enhance the topic extraction process \cite{bianchi2020cross}. They introduced a method that leverages contextualized document embeddings, resulting in improved topic quality and coherence. The study demonstrated that contextualized word embeddings produce more meaningful and coherent topic representations.

Researchers have also used hybrid approaches in recent years, leading to remarkable improvements in topic extraction.  Grootendors (2022) and  Zhang et al. (2022) adopt an innovative approach that combines TF-IDF and word embeddings \cite{grootendorst2022bertopic}, \cite{zhang2022neural}. This hybrid model uses BERT embeddings to group documents into distinct clusters and extract coherent and meaningful topics from each cluster based on TF-IDF scores. 

The model proposed in this paper enhances the topic modeling process by leveraging contextual information from SBERT embeddings \cite{reimers2019sentence} of candidate topic words within each cluster \cite{grootendorst2022bertopic}. Our new technique leverages an end-to-end semantic-driven approach using Sentence-BERT \cite{reimers2019sentence,kolesnikovadetecting} to generate better topic representations, outperforming TF-IDF, probabilistic, and other methods. This results in more coherent and meaningful topics for each cluster.

\section{Model Architecture}
The model we introduce has four modules: embedding, dimension reduction, clustering, and topic extraction. 
\begin{figure}[h] 
\centering
{\includegraphics [width=0.9  \textwidth ]{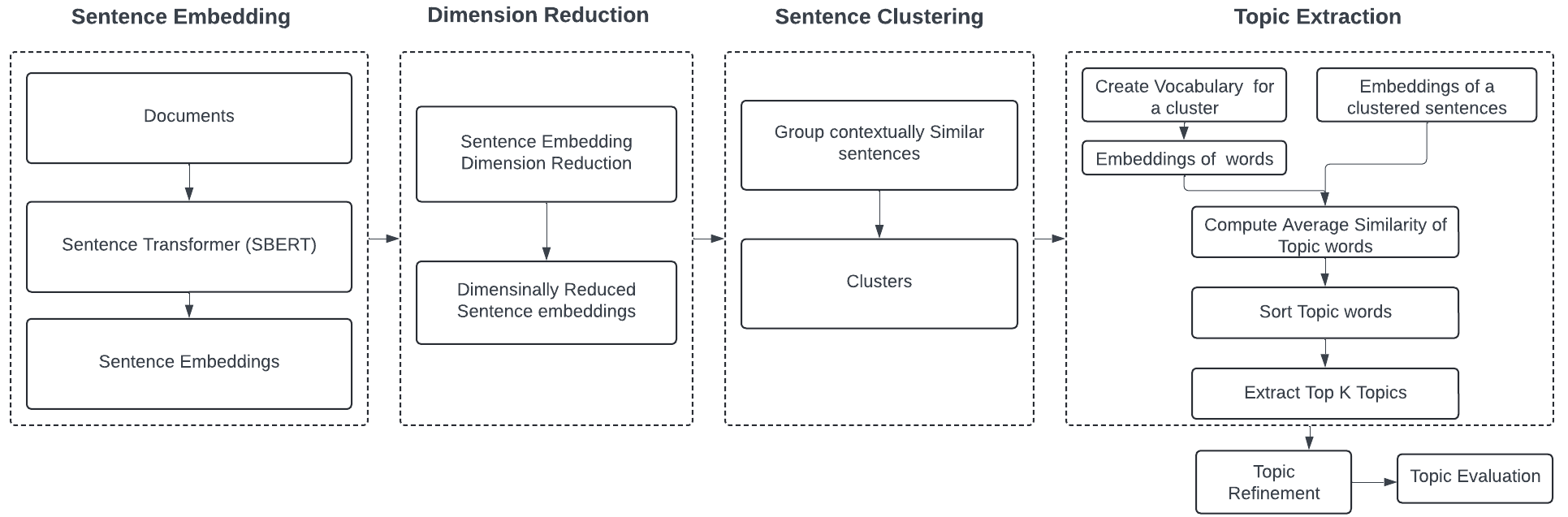}}
\caption{Overview of the proposed pipeline model architecture.}
\label{fig:pma}
\end{figure}

\subsection{Document Embedding}
In this paper, a document refers to a unit of text that can be any piece of textual content ranging from a single phrase, sentence, paragraph, or a collection of these text units or documents. The initial task in the model is creating a sentence-level vector space representation. SentenceTransformer-BERT (SBERT) \cite{reimers2019sentence,yigezuodio} is used for this purpose. SBERT converts collections of documents into high-quality sentence embeddings in a dense vector space by leveraging the BERT pre-trained language model \cite{devlin2018bert}, which provides fixed-length vector representations. In this module, any other document embedding method can be employed if it produces better vector representations and improves the quality of document clustering.  Since the clustering quality will improve as new and enhanced language models continue to emerge, the performance of the model will also improve; it is a potential benefit of our model. 

\subsection{Dimension Reduction}
Studies have shown that the proximity to the nearest data point tends to approach the distance to the farthest data point when the dimensionality of data increases \cite{aggarwal2001surprising}. As a result, the hypothesis of spatial locality becomes poorly defined in high-dimensional space, leading to diminished differences between different distance measures. This high-dimensional Sentence BERT vector space representation may challenge clustering algorithms \cite{pandove2018systematic}. Therefore, applying dimension reduction techniques is the straightforward solution for this high-dimensionality challenge to get a better clustering result \cite{allaoui2020considerably}. We employed UMAP as a dimension reduction technique that shows remarkable improvements in clustering documents, providing a significant milestone for the overall topic extraction process \cite{grootendorst2022bertopic}. We adjust UMAP's parameters, such as the number of neighbors and minimum distance, to balance the preservation of global and local structures. Furthermore, using some model explainability techniques may help to interpret UMAP output \cite{mersha2024explainable}, which is not done in this study.

\subsection{ Document Clustering}
Clustering is essential in our topic extraction process. We use reduced document embeddings, clustered based on semantic similarity, to identify and extract coherent and unique topics from a document collection. HDBSCAN is chosen for its robustness, scalability, and ability to find clusters of varying densities  \cite{mcinnes2018umap}. This method is particularly effective for diverse document structures and noisy data, providing hierarchical insights to uncover hidden topics and subtopics across the entire collection.

\subsection{Topic Extraction} 
 Topic modeling studies have demonstrated that the documents within a cluster exhibit a clear association with a specific topic \cite{grootendorst2022bertopic}. However, it is essential to realize that the documents within a cluster may contain multiple topics and subtopics, indicating a certain level of topic diversity within clusters.  Once the HDBSCAN clustering algorithm is applied and clusters are identified, the next step is detecting topic words for each cluster, building a vocabulary, and extracting topics, which involves a few steps.
First, to build a vocabulary for each cluster, sentences within each cluster are split into individual words, and these words are mapped to their corresponding contextual embedding values, helping eliminate topic-representative words that do not have any semantic contribution to the sentence.
Secondly, unique candidate words are extracted from each sentence, and an independent vocabulary is constructed for each cluster. Subsequently, contextually non-relevant unique words are eliminated from each vocabulary, resulting in a vocabulary composed of unique words associated with their embeddings.
In the third step, the average semantic similarity of each unique word within the cluster is computed by comparing it with each sentence's semantic information. This process provides an average of representative semantic similarity values for each topic word in that cluster (Equation 1). A cluster consists of a collection of $n$ unique words, represented as vocabulary $W$, accompanied by a set of $N$ contextually similar sentences denoted as $S$. To determine the representativeness of each word within the cluster, we calculate the average similarity between each word and all the sentences in the cluster; we can use  cosine/Jaccard/Euclidea similarity measurement, defined by:
\begin{equation}
\text{ave cos sim}( \vec{w_i} ) = \frac{1}{N} \sum_{j=1}^{N} \cos( \vec{w_i}, \vec{s_j} )
\end{equation}
where, $\vec{w_i}$ is the embedding vector of the $i^{th}$ word in the vocabulary $W$ and 
    $\vec{s_j}$ is the embedding vector of the $j^{th}$ sentence in the set $S$.\\
The candidate topic words are organized and sorted based on the average semantic similarity values. The top $k$ words are selected from each cluster. This process enables the extraction of topics from each cluster with enhanced accuracy and relevance of topic words specific to that cluster. After the topics are extracted, it is essential to consider how much each topic differs from others. Hence, we merge the least ranked topic with its most similar counterparts through an iterative process using similarity measures.  This iterative process helps reduce the number of topics to a user-specified value. Algorithm 1 presents a high-level overview of our model.

  \begin{algorithm}
    \caption{Topic Extraction}
    {\small
    \begin{algorithmic}[1]
        \State \textbf{Input:} Documents
        \State Create sentence embeddings
        \State Reduce sentence embedding dimensions
        \State Create clusters
        \For {cluster = 1, 2, ..., C} /* C is total number of cluster
        \State Preprocess each cluster
        \State Build a vocabulary
        \State Create word embeddings list
        \For {word = 1, 2, ..., W} /* W is total number of words in the vocabulary
        \For {sentence = 1, 2, ..., S} /* S is total number of sentences in the cluster
        \State Compute ave\_cos\_sim $w_i$ with $s_i$        
        \State $/* w_i\ is\ words\ in\ a\ cluster$
        \State $/* s_i\ is\ sentences\ in\ a\ cluster$
        \State Store the words with score values
        \EndFor
        \State Sort words
        \State Choose top $k$ words
        \State Return chosen top $k$ words        
        \EndFor        
        \EndFor
       \State $\text{mergedTopics} \gets \emptyset$
       \For {t$_i$ in topics} $\  \    \  /*\ t_i\ and\ t_j\ are\ topics $
       \For {t$_j$ in topics}
       \State If \( t_i \neq t_j \) and \( t_i \) or \( t_j \) is not merged
         
           \State simScore = computeSim(t$_i$, t$_j$)      
            \If {simScore > threshold}
              \State  newTopic = merge(t$_i$, t$_j$)            
              \State tag t$_i$ and t$_j$ as merged
               \State add newTopic to mergedTopics 
                 \EndIf
            \EndFor        
        \EndFor
\For {topic in topics}
    \If {topic is not tagged as merged}
       \State add topic to mergedTopics
       \EndIf
       \EndFor     
\State return mergedTopics
    \end{algorithmic}
    }
\end{algorithm}

\section{Experiments and Results}
In this section, we briefly discuss the experimental setup, including details about the dataset and preprocessing procedures, the model evaluation metrics employed, the performance and results of our proposed model, and the results of various model comparisons.
\subsection{Experiment setup}
We used all-MiniLML6-v2 (MiniLM) and all-mpnet-base-v2 (MPNET), two different SBERT models, in the experiments to encode documents \cite{reimers2019sentence}.
OCTIS (Optimizing and Comparing Topic Models is Simple) is an open-source Python package designed to help optimize and compare topic models \cite{terragni2021octis,yigezu2023habeshab}. It comprises a suite of tools and metrics, including topic coherence. We utilized OCTIS to conduct the model comparison experiment and validation process.

\subsection{Datasets}
The 20NewsGroups, BBC News, and Trump’s tweets datasets are used to validate our model. The 20NewsGroups dataset comprises 16,309 news articles categorized into 20 different groups \cite{lang1995newsweeder}.  The BBC News dataset contains 2,225 documents, categorized into four distinct classes, from the BBC News website between 2004 and 2005 \cite{greene2006practical}.  The 20newsgroup and BBC News datasets are a collection of short and long texts.  We used Trump’s tweets to represent more recent and short textual data \cite{grootendorst2022bertopic}. Trump's collection of tweets contains 44,253 tweets between 2009 and 2021. All these datasets are retrieved from the Kaggle repository.

\subsection{Model Evaluation}
Widely accepted and easily computable topic coherence measures, such as \(C_V\), \(C_{npmi}\), \(U_{Mass}\), and \(C_{uci}\), are used to evaluate the interpretability of topics. \\
1) \textbf{C\_V Coherence:} The C\_V coherence metric evaluates the coherence and interpretability of topics based on context vectors instead of relying on the co-occurrence frequency of words \cite{roder2015exploring}. These context vectors calculate the Normalized Pointwise Mutual Information (NPMI) between a chosen word and the frequency counts of the top topic words within the vector. The C\_V topic coherence measure correlates well with human judgment \cite{roder2015exploring}. A C\_V score of 1 indicates perfect coherence, whereas 0 indicates no coherence.\\
2) \textbf{C\_npmi:} C\_npmi (Normalized Pointwise Mutual Information coherence) works by analyzing the semantic relationships between words within a topic \cite{aletras2013evaluating}. It computes NPMI between pairs of words in each topic, measuring how strongly they are correlated with each other. C\_npmi overcomes the limitation of C\_uci by replacing PMI with normalized PMI.
 The C\_npmi measure correlates better with human judgment \cite{lau2014machine}.

 C\_npmi scores typically range from -1 to 1, where a score of 1 indicates perfect coherence.
 
C\_uci \cite{lau2014machine} and U\_mass \cite{mimno2011optimizing} measure topic coherence by observing how topic words co-occur within a topic in a reference corpus of text data. They do not depend on any other word embeddings or complex statistics like C\_npmi and C\_V. High C\_uci and U\_Mass scores indicate that the words within a topic are more coherent and have a higher likelihood of co-occurring together. 

We computed the coherence of each topic separately, and each cluster-based topic showed an excellent coherence score. These individual scores indicate that the top $k$ words in each topic have a stronger semantic relationship and a high probability of co-occurring within the given topic's context. The overall topic coherence score is computed by averaging these individual topic coherence scores.  Topic Coherence (TC) is computed for each topic model, varying the number of topics from 10 to 50 with increments of 10. We averaged the outputs from three separate runs at each interval to enhance consistency, resulting in an average score derived from a cumulative total of 15 distinct runs using fixed parameters for HDBSCAN and UMAP. Table ~\ref{tbl:metrics result} shows the four evaluation metric results.

\subsection{Model Comparison}
We compare our model with the existing traditional topic modeling approaches and ChatGPT.
\subsubsection{Traditional Models}
We conduct extensive performance comparisons between our proposed model and well-known, established models, including (LDA) \cite{blei2003latent} Latent Dirichlet Allocation, (CTM) Correlated Topic Model \cite{blei2006correlated},  ETM (Topic Modeling in Embedding Spaces) \cite{dieng2020topic}, and BERTopic \cite{grootendorst2022bertopic}. 

Topic Coherence is computed for each topic model, varying the number of topics from 10 to 50 with increments of 10. We averaged the outputs from three separate runs at each interval to enhance consistency, resulting in an average score derived from a cumulative total of 15 distinct runs. Table~\ref{tbl:table2} shows the model comparison results.
\begin{table}[t] 
 \small
\centering
\renewcommand{\arraystretch}{1.3}

\begin{minipage}{.5\textwidth}
\centering
\begin{tabular}{ p{1.3cm}|p{2.1cm}|p{1.2cm}|p{1.1cm}  }
 \hline
 \multicolumn{4}{c} {\textbf{Datasets} }\\
 \hline
 
\textbf{Metrics}& \textbf{20news group} &\textbf{BBC News}&\textbf{Trump}\\
 \hline
 C\_V& 0.735    &0.651 &  0.594 \\
 C\_npmi&0.211     &0.191    &0.205\\
 U\_mass&9.34 &8.78 &  7.94\\
 C\_uci&0.401 & 0.376& 0.322\\ 
  \hline
\end{tabular}
\caption{Topic coherence scores obtained using different model evaluation metrics using our approach. }
\label{tbl:metrics result}
\end{minipage}%
\hfill
\begin{minipage}{.47\textwidth}
\centering
\renewcommand{\arraystretch}{1.3}
\begin{tabular}{ p{2.7cm}|p{1.2cm}|p{1.2cm} }
 \hline
 \multicolumn{3}{c} {\textbf{20 Newsgroup Dataset} } \\
 \hline
 
\textbf{Models (years)}& \textbf{(C\_V)} & \textbf{(C\_npmi)}\\
 \hline
 LDA (2003)  &0.459    &0.056\\ 
 CTM (2006) &0.538		&0.042\\
 ETM (2020) &0.525		&0.095\\
 BERTopic (2022) &0.593		&0.170\\
 Our Model & \textbf{0.735}		& \textbf{0.211}\\
 
 \hline

\end{tabular}
\caption{Model comparison with C\_V and C\_npmi topic coherence metrics results}
\label{tbl:table2}
\end{minipage}
\end{table}

\subsubsection{ChatGPT}
GPT, developed for various NLP tasks such as translation, language processing, and question-answering, is described in \cite{radford2018improving}. While GPT is not explicitly designed for topic modeling and lacks integrated topic modeling algorithms, ChatGPT can generate topics and explanations by leveraging the rich information base in its embedding space. We conducted extensive experiments through programming and conversation to compare our model with ChatGPT. We split a large dataset into smaller chunks to overcome the token limit, resulting in other challenges. First, we lose critical latent themes and patterns in the document. Second, ChatGPT is stateless; it does not remember past API interactions for each chunk, particularly in multi-turn conversations, and it is difficult to process sequential data. We broke down a similar section of the 20 newsgroup datasets into chunks and extracted one topic from each chunk (Table ~\ref{fig:chunks from 20 newsgroup}). The topics generated in each chunk may not provide document-wise hidden themes and patterns. ChatGPT does not use any evaluation metrics like topic coherence and topic diversity to assess topic quality. ChatGPT generates granular topics that may need merging or splitting, but it lacks this capability. Our model allows easy topic refinement through adjustable parameters and hyperparameters.

\begin{table*}[hbt!]
\small
\centering 
{\begin{tabular}{ p{0.08\textwidth} p{12cm}}
\hline

\textbf{Chunks} & \textbf{\   \      \   Topic words}  \\\hline
Chunk 1 &JPEG, software, conversion, display, color, compression, JFIF, hardware, format\\
Chunk 2 &JPEG, GIF, Quantization, Colors, Display, Image, Quality, Hardware, Palette, Lossiness\\
Chunk 3 &JPEG, GIF, colors, quantization, display, hardware, image, palette, conversion, quality\\ 
Chunk 4 &JPEG, Compression, Huffman, Arithmetic, Coding, File, Format, Header, Quality, Data\\ 
Chunk 5 &JPEG, Compression, Decompression, Quality, Error, GIF, Conversion, Image, Degradation, Format\\
 \hline
\end{tabular}} 
\caption{One topic in each chunk with top 10 words, chunks from 20 newsgroup datasets.}
\label{fig:chunks from 20 newsgroup}
\end{table*}

Our experiments revealed that while ChatGPT performs adequately for small-size input texts, it falls short for large datasets and measuring topic quality and scalability. Compared to our models, it lacks reliability, extendability, and security for sensitive information. These limitations highlight the importance of traditional algorithms and ChatGPT and the need for enhanced techniques in topic modeling.

\subsection{Results}
Our model consistently achieves high topic coherence scores across all datasets, as various metrics show.  The model exhibits strong coherence scores when applied to preprocessed datasets. The results are shown in Table~\ref{tbl:metrics result}. Experimental results demonstrate that our proposed model outperforms traditional and embedding-based methods, including LDA, ETM, CTM, and BERTopic. For a visual representation, Figure 1(a) displays the word embedding spaces of the input dataset in reduced dimensions. Figure 1(b) illustrates the semantic clusters within the input dataset, highlighting outliers through HDBSCAN outlier detection. Figure 1(c) presents the semantic clusters of the 20 newsgroup documents, excluding the outliers. Finally, the hidden topics are extracted from each cluster, and the top 10 words from each cluster are presented in  Table~\ref{fig:certificateformat}. 

\begin{figure}[h]
    \centering
    \begin{subfigure}[b]{0.32\textwidth}
        \centering
        \includegraphics[width=\textwidth]{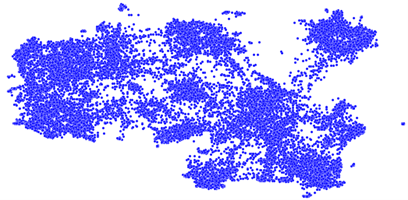}
        \caption{}
        \label{fig:image1}
    \end{subfigure}
    \hfill
    \begin{subfigure}[b]{0.25\textwidth}
        \centering
        \includegraphics[width=\textwidth]{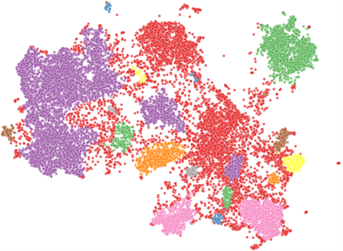}
        \caption{}
        \label{fig:image2}
    \end{subfigure}
    \hfill
    \begin{subfigure}[b]{0.25\textwidth}
        \centering
        \includegraphics[width=\textwidth]{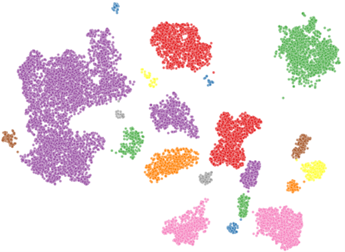}
        \caption{}
        \label{fig:image3}
    \end{subfigure}
    \caption{(a) Dimensionality reduction of 384-dimensional sentence vectors from the 20 newsgroups dataset to 2 dimensions with UMAP. (b) Highlighting semantically similar dense sentence areas via HDBSCAN clustering in dimensionally reduced sentence vectors from the 20 newsgroups dataset. Scattered red points indicate sentences labeled as noise or outliers. (c) Semantically similar dense sentence areas, excluding outlier sentences (HDBSCAN noise removal capability), were identified with HDBSCAN from the 20 newsgroups dataset.}
    \label{fig:pma}
\end{figure}

\begin{table*}[hbt!]
\small
\centering
\begin{tabular}{p{0.08\textwidth} p{0.70\textwidth} c}
\toprule
\textbf{Topic No.} & \textbf{Topic words} & \textbf{TC} \\
\midrule
1 &jesus, christ, god, bible, christians, spirit, lord, church, heaven, gospel & 0.8427 \\
2 &cars, engine, wheels, gear, brakes, tires, bike, motorcycle, parking, driving & 0.5679\\
3 &medical, health, doctor, patient, disease, cancer, symptoms, drug, physician & 0.7243\\
4 & keys, clipper, encryption, decrypt, secure, encrypted, scheme, security, algorithm& 0.7640\\
5 &beliefs, atheist, christianity, religions, atheism, christian, faith, truth, existence & 0.6008\\
6 & monitor, card, pc, disk, system, mac, scsi, window, program, display & 0.7010\\
7 &voltage, circuit, signal, resistor, diode, khz, impedance, analog, system, resistors &0.6833 \\
8 &israel, jewish, israeli, jerusalem, jews, palestinian, arab, gaza, zion, jordan & 0.7679
\\
9 &sale, price, shipping, brand, item, offer, warranty, buyer, purchased, trade & 0.6402
\\
10 &space, satellite, launch, orbit, earth, spacecraft, shuttle, moon, nasa, mission &0.5832
 \\
11 &weapon, firearm, guns, handguns, crime, laws, amendment, firearms, govern, right &0.5892
 \\
12 &season,game, teams , hockey, playoff, defenseman, goal, score, player, penalty & 0.7491
\\
13 &research, project, conference, acm, proceedings, papers, publication, journal & 0.7585
\\
14 &thanks, appreciate, reply, response, email, respond, welcome, advance, answer &0.6783
 \\
15 &bus, eisa, cards, ide, vesa, svga, isa, video, bios, motherboard &0.5695
 \\
16 &sunos, gcc, compile, lib, libraries, patch, login, window, unix, xdm & 0.7847
\\
17 &drive, ide, disk, boot, jumper, controller, floppy, tape, dma, master &0.6654
 \\
18 &window, program, file, server, user, run, version, openwindows, ftp, xview &0.5297
 \\
19 &printers, print, ink, hp, deskjet, laser, paper, printing, printer,document &0.7899 \\
20 &law, govern, protect, legal, citizen, right, policy, control, crime, people &0.7104 \\

\midrule
& \textbf{Average Topic Coherence} & \textbf{0.6850} \\
\bottomrule
\end{tabular}

\caption{Topics, top 10 topic words, and c\_v individual topic coherence scores for 20 newsgroup datasets, with overall topic coherence score as the average of individual scores.}
\label{fig:certificateformat}
\end{table*}

\subsection{Model Performance}
Our model exhibits several notable strengths compared to the other topic models we compared with this study. The utilization of end-to-end embedding approaches for topic modeling provides many advantages to our model. First, our model is adaptable to different language models since it depends on embedding spaces for clustering, enabling it to stay at the forefront of advances in embedding techniques, ensuring its continuous upgrading and scalability in line with the latest developments in the field. Second, the most significant strength lies in cluster-based vocabulary construction and contextual similarity computation. These processes leverage the inherent contextual similarity among words and sentences within clusters, empowering the model to generate coherent and meaningful topics consistently.

\subsection{Discussion}
We have presented a novel model, an unsupervised learning algorithm designed to discover topics within a semantic space that leverages the embedding of documents. We have demonstrated how the semantic vector space is used for the representation of topics, enabling the computation of topics by identifying dense regions of highly semantically similar documents. To understand our model comprehensively, it is essential to understand the contextual importance of each word within a document and sentence from the Transformer model. The model centers on each word's and sentence's contextual meaning and contribution within its corresponding cluster or semantic space. These central concepts offer two main advantages to the model. Firstly, we employ Sentence-Transformer's word embedding values to extract topics based on the relevance of each word within its cluster using some similarity measure. Secondly, we exclude non-relevant words from the topic extraction process by utilizing similarity score values, enhancing the model's performance. 
 HDBSCAN identifies highly semantically similar dense and sparse sentence areas in the sentence vector space on the UMAP dimensionally reduced sentence vector. Those semantically similar dense areas are where we are interested in finding the underlying topics. In our finding, sparse sentence areas are semantically less similar to each other and also to the dense sentence areas, as shown in Fig  1(b).  These sparse areas are considered as noise, and no significant underlying topic exists, and we exclude them from the topic extraction process, as shown in Fig 1(c). The $ minimum\ cluster\ size $ is the most critical hyperparameter in HDBSCAN. In our experiments, we determined that a $minimum\ cluster\ size $ of 10 returns the best results for 20 newsgroup and BBCNews datasets and 8 for Trump's Twitt dataset. We notice that larger values increase the likelihood of merging unrelated sentence clusters. Using cosine similarity, we computed the topics for each identified dense area or cluster. Topics exhibiting high cosine similarity values, indicating close to 1, are considered highly similar. Depending on the desired level of reduction, the users can set a threshold similarity score for the user-specified values to their preferences. For example in Table ~\ref{fig:certificateformat}, we can merge topics 7 and 17 into the 'hardware' category, and merge topics 16 and 18 into the 'software' category.

\subsection{Limitation of the Study}
Traditional topic modeling techniques depend on the frequency of words. Our semantic-driven topic modeling technique focuses on the meaning of words and documents instead of their surface characteristics, which is our study's greatest strength and new paradigm shift in the topic modeling study. Our model has a limitation in detecting latent subtopics. Latent subtopics are topics that are not directly stated but are suggested. For example, consider the customer feedback about the Apple Smartphone and the model identified explicit topics such as camera quality, screen size, battery life, storage, and processing speed. However, our model does not detect latent subtopics like the user's overall satisfaction. This subtopic identification is a common challenge for many topic modeling techniques and is an open research area.
\section{Conclusion}
We have introduced a novel approach to topic modeling that leverages the rich contextual information provided by transformer models to generate topics from a collection of documents. The model employs the SBERT to obtain sentence embeddings, reduces the dimensions of these sentence embeddings, identifies semantically similar dense sentence vector spaces using a density-based clustering algorithm, and extracts coherent topics that represent these semantically dense areas or clusters. Our experiments have shown that the proposed model achieves competitive results and performance compared to various existing models across different datasets.

\bibliographystyle{elsarticle-num} 
 \bibliography{cas-refs}

\begin{thebibliography}{10}
\expandafter\ifx\csname url\endcsname\relax
  \def\url#1{\texttt{#1}}\fi
\expandafter\ifx\csname urlprefix\endcsname\relax\def\urlprefix{URL }\fi
\expandafter\ifx\csname href\endcsname\relax
  \def\href#1#2{#2} \def\path#1{#1}\fi

\bibitem{blei2012probabilistic}
D.~M. Blei, Probabilistic topic models, Communications of the ACM 55~(4) (2012) 77--84.

\bibitem{rodriguez2020computational}
M.~Y. Rodriguez, H.~Storer, A computational social science perspective on qualitative data exploration: Using topic models for the descriptive analysis of social media data, Journal of Technology in Human Services 38~(1) (2020) 54--86.

\bibitem{joshi2023deepsumm}
A.~Joshi, E.~Fidalgo, E.~Alegre, L.~Fern{\'a}ndez-Robles, Deepsumm: Exploiting topic models and sequence to sequence networks for extractive text summarization, Expert Systems with Applications 211 (2023) 118442.

\bibitem{albalawi2020using}
R.~Albalawi, T.~H. Yeap, M.~Benyoucef, Using topic modeling methods for short-text data: A comparative analysis, Frontiers in Artificial Intelligence 3 (2020) 42.

\bibitem{punziano2023digital}
G.~Punziano, C.~C. De~Falco, D.~Trezza, Digital mixed content analysis for the study of digital platform social data: An illustration from the analysis of covid-19 risk perception in the italian twittersphere, Journal of Mixed Methods Research 17~(2) (2023) 143--170.

\bibitem{mustak2021artificial}
M.~Mustak, J.~Salminen, L.~Pl{\'e}, J.~Wirtz, Artificial intelligence in marketing: Topic modeling, scientometric analysis, and research agenda, Journal of Business Research 124 (2021) 389--404.

\bibitem{sbalchiero2020topic}
S.~Sbalchiero, M.~Eder, Topic modeling, long texts and the best number of topics. some problems and solutions, Quality \& Quantity 54 (2020) 1095--1108.

\bibitem{blei2003latent}
D.~M. Blei, A.~Y. Ng, M.~I. Jordan, Latent dirichlet allocation, Journal of Machine Learning Research 3~(Jan) (2003) 993--1022.

\bibitem{fevotte2011algorithms}
C.~F{\'e}votte, J.~Idier, Algorithms for nonnegative matrix factorization with the $\beta$-divergence, Neural Computation 23~(9) (2011) 2421--2456.

\bibitem{hofmann1999probabilistic}
T.~Hofmann, Probabilistic latent semantic indexing, in: Proceedings of the 22nd Annual International ACM SIGIR Conference on Research and Development in Information Retrieval, 1999, pp. 50--57.

\bibitem{grootendorst2022bertopic}
M.~Grootendorst, Bertopic: Neural topic modeling with a class-based tf-idf procedure, arXiv preprint arXiv:2203.05794 (2022).

\bibitem{dieng2020topic}
A.~B. Dieng, F.~J. Ruiz, D.~M. Blei, Topic modeling in embedding spaces, Transactions of the Association for Computational Linguistics 8 (2020) 439--453.

\bibitem{blei2006correlated}
D.~Blei, J.~Lafferty, Correlated topic models, Advances in Neural Information Processing Systems 18 (2006) 147.

\bibitem{zhao2021topic}
H.~Zhao, D.~Phung, V.~Huynh, Y.~Jin, L.~Du, W.~Buntine, Topic modelling meets deep neural networks: A survey, arXiv preprint arXiv:2103.00498 (2021).

\bibitem{terragni2021octis}
S.~Terragni, E.~Fersini, B.~G. Galuzzi, P.~Tropeano, A.~Candelieri, Octis: Comparing and optimizing topic models is simple!, in: Proceedings of the 16th Conference of the European Chapter of the Association for Computational Linguistics: System Demonstrations, 2021, pp. 263--270.

\bibitem{agarwal2021comparative}
N.~Agarwal, G.~Sikka, L.~K. Awasthi, Comparative study of topic modeling and word embedding approaches for web service clustering, in: 2021 Thirteenth International Conference on Contemporary Computing (IC3-2021), 2021, pp. 309--313.

\bibitem{qiang2017topic}
J.~Qiang, P.~Chen, T.~Wang, X.~Wu, Topic modeling over short texts by incorporating word embeddings, in: Advances in Knowledge Discovery and Data Mining: 21st Pacific-Asia Conference, PAKDD 2017, Jeju, South Korea, May 23-26, 2017, Proceedings, Part II 21, Springer, 2017, pp. 363--374.

\bibitem{bianchi2020cross}
F.~Bianchi, S.~Terragni, D.~Hovy, D.~Nozza, E.~Fersini, Cross-lingual contextualized topic models with zero-shot learning, arXiv preprint arXiv:2004.07737 (2020).

\bibitem{yigezu2023habesha}
M.~G. Yigezu, T.~Kebede, O.~Kolesnikova, G.~Sidorov, A.~Gelbukh, Habesha@ dravidianlangtech: Utilizing deep and transfer learning approaches for sentiment analysis., in: Proceedings of the Third Workshop on Speech and Language Technologies for Dravidian Languages, 2023, pp. 239--243.

\bibitem{angelov2020top2vec}
D.~Angelov, Top2vec: Distributed representations of topics, arXiv preprint arXiv:2008.09470 (2020).

\bibitem{zhang2022neural}
Z.~Zhang, M.~Fang, L.~Chen, M.-R. Namazi-Rad, Is neural topic modelling better than clustering? an empirical study on clustering with contextual embeddings for topics, arXiv preprint arXiv:2204.09874 (2022).

\bibitem{reimers2019sentence}
N.~Reimers, I.~Gurevych, Sentence-bert: Sentence embeddings using siamese bert-networks, arXiv preprint arXiv:1908.10084 (2019).

\bibitem{kolesnikovadetecting}
O.~Kolesnikova, M.~G. Yigezu, A.~Gelbukh, S.~Abitte, G.~Sidorov, Detecting multilingual hate speech targeting immigrants and women on twitter, Journal of Intelligent \& Fuzzy Systems~(Preprint)  1--10.

\bibitem{yigezuodio}
M.~G. Yigezu, O.~Kolesnikova, A.~Gelbukh, G.~Sidorov, Odio-bert: Evaluating domain task impact in hate speech detection, Journal of Intelligent \& Fuzzy Systems~(Preprint)  1--12.

\bibitem{devlin2018bert}
J.~Devlin, M.-W. Chang, K.~Lee, K.~Toutanova, Bert: Pre-training of deep bidirectional transformers for language understanding, arXiv preprint arXiv:1810.04805 (2018).

\bibitem{aggarwal2001surprising}
C.~C. Aggarwal, A.~Hinneburg, D.~A. Keim, On the surprising behavior of distance metrics in high dimensional space, in: Database Theory—ICDT 2001: 8th International Conference London, UK, January 4--6, 2001 Proceedings 8, Springer, 2001, pp. 420--434.

\bibitem{pandove2018systematic}
D.~Pandove, S.~Goel, R.~Rani, Systematic review of clustering high-dimensional and large datasets, ACM Transactions on Knowledge Discovery from Data (TKDD) 12~(2) (2018) 1--68.

\bibitem{allaoui2020considerably}
M.~Allaoui, M.~L. Kherfi, A.~Cheriet, Considerably improving clustering algorithms using umap dimensionality reduction technique: A comparative study, in: International conference on image and signal processing, Springer, 2020, pp. 317--325.

\bibitem{mersha2024explainable}
M.~Mersha, K.~Lam, J.~Wood, A.~AlShami, J.~Kalita, Explainable artificial intelligence: A survey of needs, techniques, applications, and future direction, Neurocomputing (2024) 128111.

\bibitem{mcinnes2018umap}
L.~McInnes, J.~Healy, J.~Melville, Umap: Uniform manifold approximation and projection for dimension reduction, arXiv preprint arXiv:1802.03426 (2018).

\bibitem{yigezu2023habeshab}
M.~G. Yigezu, S.~Kanta, O.~Kolesnikova, G.~Sidorov, A.~Gelbukh, Habesha@ dravidianlangtech: Abusive comment detection using deep learning approach, in: Proceedings of the Third Workshop on Speech and Language Technologies for Dravidian Languages, 2023, pp. 244--249.

\bibitem{lang1995newsweeder}
K.~Lang, Newsweeder: Learning to filter netnews, in: Machine learning proceedings 1995, Elsevier, 1995, pp. 331--339.

\bibitem{greene2006practical}
D.~Greene, P.~Cunningham, Practical solutions to the problem of diagonal dominance in kernel document clustering, in: Proceedings of the 23rd International Conference on Machine learning, 2006, pp. 377--384.

\bibitem{roder2015exploring}
M.~R{\"o}der, A.~Both, A.~Hinneburg, Exploring the space of topic coherence measures, in: Proceedings of the Eighth ACM International Conference on Web Search and Data Mining, 2015, pp. 399--408.

\bibitem{aletras2013evaluating}
N.~Aletras, M.~Stevenson, Evaluating topic coherence using distributional semantics, in: Proceedings of the 10th International Conference on Computational Semantics (IWCS 2013)--Long Papers, 2013, pp. 13--22.

\bibitem{lau2014machine}
J.~H. Lau, D.~Newman, T.~Baldwin, Machine reading tea leaves: Automatically evaluating topic coherence and topic model quality, in: Proceedings of the 14th Conference of the European Chapter of the Association for Computational Linguistics, 2014, pp. 530--539.

\bibitem{mimno2011optimizing}
D.~Mimno, H.~Wallach, E.~Talley, M.~Leenders, A.~McCallum, Optimizing semantic coherence in topic models, in: Proceedings of the 2011 Conference on Empirical Methods in Natural Language Processing, 2011, pp. 262--272.

\bibitem{radford2018improving}
A.~Radford, K.~Narasimhan, T.~Salimans, I.~Sutskever, et~al., Improving language understanding by generative pre-training (2018).

\end{thebibliography}

\appendix

\section{ BBC News and Trump’s Tweet datasets Results }
\label{sec:sample:appendix}
\begin{table*}[hbt!]
\small
\centering
\begin{tabular}{p{0.08\textwidth} p{0.70\textwidth} c}
\toprule
\textbf{Topic No.} & \textbf{Topic words} & \textbf{TC} \\
\midrule
1 & government, policy, election, prime, minister, parliament, party, vote, campaign, leader & 0.7106 \\
2 & market, stock, investment, company, profit, share, growth, trade, financial, economic & 0.6498 \\
3 & technology, innovation, software, hardware, device, internet, application, development, computer, AI & 0.8891 \\
4 & sport, match, team, player, coach, tournament, championship, league, score, goal & 0.8856 \\
5 & film, movie, actor, director, production, release, cinema, audience, award, genre & 0.5959 \\
6 & music, album, artist, song, concert, band, release, genre, chart, festival & 0.7686 \\
7 & healthcare, hospital, doctor, patient, treatment, disease, research, vaccine, medicine, clinic & 0.7780 \\
8 & education, school, student, university, teacher, curriculum, learning, exam, degree, research & 0.7048 \\
9 & finance, banking, interest, loan, credit, debt, mortgage, investment, rate, account & 0.7113 \\
10 & travel, destination, tourism, flight, hotel, vacation, trip, itinerary, tourist, booking & 0.7202 \\
11 & environment, climate, pollution, conservation, wildlife, sustainability, energy, emission, ecosystem, habitat & 0.6290 \\
12 & economy, growth, recession, inflation, employment, market, GDP, sector, trade, investment & 0.5249 \\
13 & fashion, design, trend, style, collection, brand, runway, model, fabric, accessory & 0.5888 \\
14 & science, research, discovery, experiment, theory, laboratory, innovation, technology, study, data & 0.6870 \\
15 & space, planet, mission, satellite, NASA, astronomy, galaxy, launch, exploration, rocket & 0.6750 \\
16 & law, court, legal, case, judge, lawyer, trial, justice, verdict, crime & 0.7174 \\
17 & politics, election, candidate, debate, policy, government, vote, campaign, party, issue & 0.6904 \\
18 & culture, tradition, festival, heritage, community, art, history, celebration, custom, belief & 0.8849 \\
19 & social, media, platform, network, content, user, engagement, post, trend, digital & 0.7712 \\
20 & automotive, car, vehicle, engine, model, manufacturer, technology, design, performance, fuel & 0.6570 \\
\midrule
& \textbf{Average Topic Coherence} & \textbf{0.7150} \\
\bottomrule
\end{tabular}
\caption{Topics, top 10 topic words, and c\_v individual topic coherence scores for BBC News datasets, with overall topic coherence score as the average of individual scores.}
\end{table*}

\begin{table*}[hbt!]
\small
\centering
\begin{tabular}{p{0.08\textwidth} p{0.70\textwidth} c}
\toprule
\textbf{Topic No.} & \textbf{Topic words} & \textbf{TC} \\
\midrule
1 & campaign, election, vote, win, rally, support, candidate, primaries, poll, turnout & 0.5288 \\
2 & economy, jobs, growth, market, trade, stock, business, investment, manufacturing, economic & 0.4969 \\
3 & media, news, journalist, report, coverage, CNN, NYTimes, article, bias, truth & 0.5562 \\
4 & America, great, country, patriotism, citizens, USA, nation, flag, independence, freedom & 0.7484 \\
5 & security, border, immigration, wall, illegal, ICE, enforcement, policy, crime, safety & 0.7573 \\
6 & military, troops, veterans, defense, service, army, navy, honor, sacrifice, support & 0.6159 \\
7 & healthcare, Obamacare, insurance, policy, reform, prescription, cost, doctors, patients, coverage & 0.6844 \\
8 & law, justice, court, judge, trial, legal, crime, investigation, verdict, FBI & 0.5192 \\
9 & foreign policy, trade, China, tariffs, agreement, negotiation, allies, relations, diplomacy, sanctions & 0.5791 \\
10 & tax, reform, cuts, policy, income, IRS, corporate, middle class, reduction, plan & 0.6121 \\
11 & energy, oil, gas, production, pipeline, industry, policy, prices, renewable, coal & 0.7001 \\
12 & education, schools, students, teachers, policy, funding, reform, curriculum, learning, college & 0.4468 \\
13 & impeachment, investigation, trial, defense, Democrats, hearing, testimony, witnesses, charges, inquiry & 0.8329 \\
14 & COVID-19, pandemic, virus, vaccine, response, cases, testing, treatment, healthcare, guidelines & 0.7353 \\
15 & Second Amendment, rights, firearms, NRA, legislation, ownership, control, safety, defense, law & 0.6174 \\
16 & tweets, retweets, followers, media, platform, engagement, post, message, hashtag, account & 0.5459 \\
17 & infrastructure, projects, development, funding, roads, bridges, construction, transportation, investment, plan & 0.6704 \\
18 & trade, negotiation, NAFTA, agreement, USMCA, exports, imports, tariffs, balance, partners & 0.5642 \\
19 & climate, environment, policy, Paris, emissions, energy, sustainability, conservation, regulation, impact & 0.7653 \\
20 & elections, fraud, recount, integrity, ballots, results, dispute, claims, process, certification & 0.6879 \\
\midrule
& \textbf{   Average Topic Coherence} & \textbf{0.6332} \\
\bottomrule
\end{tabular}
\caption{Topics, top 10 topic words, and c\_v individual topic coherence scores for Trump's Tweet datasets, with overall topic coherence score as the average of individual scores.}
\end{table*}






\end{document}